\documentclass{article}

\usepackage[preprint, nonatbib]{neurips_2025} 
\usepackage[numbers]{natbib}

\usepackage[utf8]{inputenc} 
\usepackage[T1]{fontenc}    
\usepackage[hidelinks]{hyperref}       
\usepackage{url}            
\usepackage{booktabs}       
\usepackage{amsfonts}       
\usepackage{nicefrac}       
\usepackage{microtype}      
\usepackage{xcolor}         
\usepackage{pgfplots}       
\usepackage{pgfplotstable}
\pgfplotsset{compat=1.18} 
\usepgfplotslibrary{groupplots}
\usepackage[utf8]{inputenc}
\usepackage{framed}
\usepackage{caption}

\title{Smarter Together: Creating Agentic Communities of Practice through Shared Experiential Learning}

\author{%
  Valentin~Tablan\thanks{Main contact} \\
  The Memory Company\\
  \texttt{valentin@memco.ai} \\
  \And
  Scott~Taylor \\
  The Memory Company \\
  \And
  Gabriel~Hurtado \\
  Moonsong Labs, Inc. \\
  \And
  Kristoffer~Bernhem \\
  The Memory Company \\
  \And
  Anders~Uhrenholt \\
  Moonsong Labs, Inc. \\
  \And
  Gabriele~Farei \\
  Moonsong Labs, Inc. \\
  \And
  Karo~Moilanen \\
  Moonsong Labs, Inc. \\
}

\begin{document}

\maketitle

\begin{abstract}
    The transition from human-centric to agent-centric software development practices is disrupting existing knowledge sharing environments for software developers. Traditional peer-to-peer repositories and developer communities for shared technical knowledge and best practice have witnessed dramatic drops in participation in a short period of time. At the same time, agentic functional equivalents are yet to emerge leaving AI agents, which already generate a significant proportion of all new software code produced, without access to repositories of valuable shared learning. 
    
    In this paper, we introduce Spark, a novel shared agentic memory architecture which is designed to emulate the collective intelligence and know-how of human developer communities. Spark enables AI coding agents to both contribute to and draw from a persistent and continuously evolving experiential memory. Agents operating in the same general problem space use the Spark shared memory as a repository of new knowledge to achieve {\em collective continual learning}.
        
    We evaluate Spark as a coach for AI coding agents performing software develop\-ment tasks. We demonstrate that recommendations made by Spark improve the quality of code generated by generic code generation models at varying sizes and capability tiers. Boosted by Spark, a small open-weights model with 30 billion parameters was able to match the code quality afforded by a much larger state-of-the-art model.
    
    Separately, we measure the intrinsic quality of recommendations generated by Spark against a wide range of criteria inspired by software development best practice, and achieve helpfulness levels of up to 98.2\% in the top two (out of five) qualitative helpfulness bands.

\end{abstract}

\section{Introduction}

Over the past year, the software developer ecosystem has shifted from human-centric tooling to a paradigm best described as \textit{``human-AI collectives"}. Code agents which write, refactor, and reason about software at all levels of abstraction across all architectural areas now act as first-class participants in code editors, IDEs, CLIs, and CI/CD/CL pipelines. This rapid transformation has given rise to a structural asymmetry. While code generation models maintain their improvement trajectory in sheer raw capability, their experience and situational awareness remain necessarily localized and bound to a \textit{specific} execution run, project, user, or provider. Due to the lack of continual learning, each interaction starts a new learning loop which has a limited lifespan, and the combined effect of such silos and amnesia puts a significant limit on the utility of AI coding agents.

In order to enable genuine continual learning, and to let AI agents acquire new skills \textit{``on the job"}, either online changes to the weights of the underlying foundation models, or some form of external memory are called for. Our approach falls in the latter paradigm of continual learning via external memory. Existing implementations of external memory for agents are designed with individual users in mind. They are made available by providers of upstream foundation models or take the form of a separate add-on which the user is required to integrate with a given workflow and toolchain. The former introduces a risk of lock-in to a particular provider, while the latter places the onus on the developer to integrate, optimize, and maintain a given memory implementation.

We observe that human creativity is often collective rather than individual, with communities of practice emerging from practitioners who work on similar problems, and who benefit from sharing, maintaining, and cultivating know\-ledge and domain expertise. Based on this insight, we propose a {\em shared agentic memory} layer, a space in which multiple agents jointly contribute, develop, and refine knowledge, and benefit from its continuous accretion without silos or amnesia.

Our contribution is two-fold: we (i) propose a community memory architecture as an alternative to closed, siloed, single-user memory topologies, and (ii) measure its utility in a configuration and use case aimed at agents that support code generation as well as optimizing developer productivity and experience. We believe that open, shared memory is generalizable across domains, and have chosen coding agents as our first implementation domain due to their traction and widespread use in the market.

\subsection{Spark}

Current state-of-the-art AI agents are built on top of large language models (LLMs) which are unlikely to be updated and retrained during the lifetime of a typical agent. Such asynchrony axiomatically limits agents'~ability to learn new skills and behaviors continuously and over long time horizons. Even when agents are able to retain information within their context window at a given point in time, crucial information is (i) eventually dropped from the context of the current session, (ii) typically not shared between sessions, and (iii) not shared between agents working with different users. Furthermore, existing memory extensions to agents represent within-context information as raw static snapshot records which do not evolve, very much continuing in the tradition of the mechanical store-and-recall persistence mode associated with traditional databases. The absence of a shared long-term memory which is able to reflect agents' continuous experiences is particularly pronounced and detrimental in the domain of software development and the emerging development paradigms of agentic coding and~\textit{``vibe coding"} (\citep{sapkota2025vibecodingvsagentic}).

We introduce {\em shared agentic memory}: memory used collectively by a group of AI agents operating in the same broad task domain. Spark is the first instantiation of that concept, applied to agents that support software development. Spark treats every coding agent interaction as a potential learning event which is (i) captured as a structured and portable unit (trace) of experience, (ii) merged into a collective knowledge space through a separate continuous and autonomous curation process, and (iii) redistributed to coding agents that face similar challenges in the future. By doing so, Spark operationalizes collective learning and restores the self-healing and self-optimizing property that developer communities \textit{had} before Generative AI became an intermediary controller of how knowledge, expertise, and practical know-how exists, spreads, and evolves among developers.

\subsection{Paper Summary}

\begin{figure}[h!]
    \begin{center}
        \begin{tikzpicture}
            \definecolor{colorNoSpark}{RGB}{128,128,128}   
            \definecolor{colorSparkDoc}{RGB}{50,150,50}    
            \definecolor{colorSparkDocExp}{RGB}{255,127,0}     

            \begin{axis}[
                ybar=0pt,
                bar width=8pt,
                ymin=4.1,
                ymax=5,
                ylabel={Mean Code Quality},
                enlarge x limits={0.15},
                axis lines*=left,
                grid=major,
                ytick={4, 4.2, 4.4, 4.6, 4.8, 5},
                major grid style={dotted, gray!50},
                error bars/y dir=both,                
                error bars/y explicit,                
                error bars/error bar style={line width=0.5pt, black}, 
                xtick={1, 2, 3, 4},
                xticklabels={Human, Qwen3-Coder, Haiku-4.5, GPT5-Codex},
                xtick pos=bottom,
                tick label style={font=\footnotesize},
                legend style={
                    at={(0.5,-0.2)},
                    anchor=north,
                    legend columns=-1,
                    draw=none,
                    font=\footnotesize
                }
            ]
            \pgfmathsetmacro{\offsetA}{0} 
            \pgfmathsetmacro{\offsetB}{0}     
            \pgfmathsetmacro{\offsetC}{0.15}  
            \pgfplotstableread[col sep=comma]{eval_data_baseline.csv}\loadedtablebase
            
            \addplot[
                ybar,
                fill=colorNoSpark,
                draw=black,
                xshift=\offsetA cm
            ] table [
                x expr=\coordindex+1,
                y=mean_score,
                y error=std_error
            ]{\loadedtablebase};
            \addlegendentry{\texttt{NO-SPARK}\enspace};
            \pgfplotstableread[col sep=comma]{eval_data_exps.csv}\loadedtableexps
            \addplot[
                ybar,
                fill=colorSparkDocExp,
                draw=black,
                xshift=\offsetC cm
            ] table [
                x expr=\coordindex+1,
                y=mean_score,
                y error=std_error,
            ]
            {\loadedtableexps};
            \addlegendentry{\texttt{WITH-SPARK}}; 
            \end{axis}
        \end{tikzpicture}
    \end{center}
\captionsetup{format=hang}
\caption{\footnotesize \textbf{Quality of code generated by LLMs with vs. without Spark}. Quality of code produced by LLMs under the two conditions. 1) \texttt{NO-SPARK}: \textbf{baseline} performance of a given LLM backbone with \underline{no} access to an external Spark memory. 2) \texttt{WITH-SPARK}: code generation using an external Spark memory populated with raw public software \textbf{documentation} and curated knowledge extracted from synthetic \textbf{experiential traces}.\\
Code quality, judged by an independent LLM judge, is evaluated on 1000 Python data science problems from the DS-1000 data set. NB. the error bars represent standard error. The ``Human" data point represents the quality of human-provided reference solutions as per the DS-1000 data set.}
\label{fig:headline_chart}
\label{fig:quality_scores}
\end{figure}
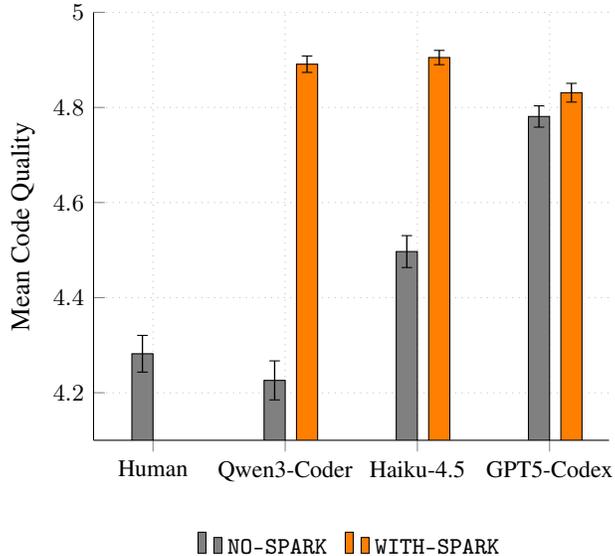

In this study, we evaluate Spark as a booster for existing memory-less coding agents. We measure the quality of the code produced by three LLMs under two distinct conditions. The \verb+NO SPARK+ baseline condition measures the raw performance of LLMs without access to any external memory. The \verb+WITH-SPARK+ condition grants LLMs access to a collective Spark memory space populated with publicly available raw software documentation and synthetic experiential traces. The LLMs used in our experiments were selected to cover a representative range of model sizes as well as both open-weights and closed source variants: \verb+Qwen3-Coder-30B-A3B-Instruct+ (small, open-weights), \verb+Anthropic Haiku 4.5+ (medium, commercial), and \verb+GPT5-Codex+ (large, state of the art, commercial).

Our qualitative measurements were made by \verb+Gemini 2.5 Pro+ acting as an independent judge using a qualitative 1-5 score for code solution completeness and adherence to good software development practices. The choice of the judge LLM was motivated by its strength and independence in relation to other LLM providers used in our study to avoid spurious correlations between candidate and judge models. 

Our code quality results are summarized in Fig.~\ref{fig:headline_chart}. We further evaluate the generic helpfulness of Spark's recommendations.

\section{Related Work}
\label{related_work_section}

In this study, we focus specifically on the memory aspects of agentic and agent-like systems designed for software development and developer tooling. For generic surveys of recent memory approaches and design choices for agents, LLMs, and wider aspects of AI, see
~\citep{du2025rethinkingmemoryaitaxonomy}, ~\citep{zhang2024surveymemorymechanismlarge}, ~\citep{wu2025humanmemoryaimemory}, ~\citep{he2025humaninspiredperspectivessurveyai}, ~\citep{shan2025cognitivememorylargelanguage}, \citep{liu2025comprehensivesurveylongcontext}, and ~\citep{sumers2024cognitivearchitectureslanguageagents}.

\subsection{Single User, Provider-managed Memory}

Commercial providers of foundation models have begun to acknowledge the need for external memory as a key mechanism to supplement the generic functionality of foundation models. Recent shallow memory extensions have focused mainly on how to personalize user experience, and have typically taken the form of opaque memory stores (such as ChatGPT's memory\footnote{\url{https://openai.com/}}) or a set of Markdown files that are loaded into the model context at runtime. Regarding the latter, Anthropic's memory subsystem for Claude\footnote{\url{https://docs.claude.com/en/docs/claude-code/memory}} provides the ability to store organization-, team-, user-, and project-level memory traces in simple \texttt{.md} files. 

In recent implementations of the single-user, provider-managed memory design principle, memory traces are in some cases organized hierarchically. However, they offer no mechanisms or support for \textit{automatic and autonomous} information flows across different memory levels: for example, individual user-level agents are not able to contribute to the shared population-level knowledge of a team or an organization. From an operational point of view, memory instances of this type are cumbersome because they presuppose that the user, the team, and the organization manage all of their memory records. They must do so using whichever existing tools or proprietary processes they have at their disposal, without the benefit of any provider-supplied end-to-end tooling covering all aspects of memory management. 

The single-user, provider-managed memory design is inherently closed and siloed, and heavily constrains code agents' learning capabilities across developer communities.

\subsection{Single User, User-managed Memory}

Beyond LLM providers' basic closed memory extensions, a range of slightly more substantial memory implementations offer facilities and tools for users to manage their own memory stores, independently of upstream providers. Recent tools of this kind similarly rely on single-user memory stores, but also provide some tools for memory management beyond simple editing of \texttt{.md} files.

Notable recent examples include Letta\footnote{\url{https://www.letta.com/}} which creates a memory layer for stateful agents, and Zep\footnote{\url{{https://www.getzep.com/product/agent-memory/}}}~\citep{rasmussen2025zeptemporalknowledgegraph}, a memory layer for agents with a knowledge graph for entities and facts extracted from chat histories and business data. ByteRover\footnote{\url{https://www.byterover.dev/}} provides a memory layer for coding agents to represent previous interactions, context, and experiences during coding tasks. Mem0\footnote{\url{https://mem0.ai/}}~\citep{chhikara2025mem0buildingproductionreadyai} offers a memory layer for LLM applications while Cognee's\footnote{\url{https://www.cognee.ai/}} memory layer for agents uses graph and vector memory in conjunction with ontologies.

Although the single-user, user-managed memory design is somewhat more open, it is still fragmented and siloed due to the absence of true portability, true semantic interoperability, and seamless cross-platform use across developer communities spanning diverse memory representations, requirements, and tools. Recent memory services of this type still lead to an operational burden because they only cover some aspects and stages of end-to-end memory management, and because they introduce additional, compounding dependency effects when used jointly.

\subsection{Lower-level Building Blocks for Memory Mechanisms}

Other approaches have explored and provide software libraries, utilities, and resources that can form part of an end-to-end agentic memory management solution, should a developer wish to implement one from scratch. Inspired by the design principles of operating systems, MemOS~\citep{li2025memosmemoryosai} propose a generic memory-oriented framework for LLMs which combines memory representations, scheduling, and evolution with multi-modal storage and retrieval inspired by OS-like operations and concerns such as memory read, scheduling, memory flow, and access control. In the realm of lightweight framework- and library-specific memory extensions, CrewAI\footnote{\url{https://docs.crewai.com/en/concepts/memory}} provide a Python framework for building AI agents, including facilities for memory management. Memobase\footnote{\url{https://www.memobase.io/}} is a (user) profile-based memory add-on layer for agents and consumer applications. A deep research agent called Memento~\citep{zhou2025mementofinetuningllmagents} exploits a continuous memory-augmented Markov Decision Process via memory-based online reinforcement learning using a \textit{Q}-function which mimics case-based reasoning. \citep{westhäußer2025caimdevelopmentevaluationcognitive} propose a memory framework with a controller, memory retrieval, and memory storage layers. \citep{xu2025amemagenticmemoryllm} propose a memory system which exploits interconnected knowledge networks through dynamic indexing and linking inspired by the the Zettelkasten method. Similarly, \citep{salama2025meminsightautonomousmemoryaugmentation} explore a memory augmentation approach to enhance shallow-semantic data representation and retrieval mechanisms with multi-level attributes and annotations.

A variety of partial memory mechanisms act as extensions to LLM(Ops) libraries (such as LlamaIndex\footnote{\url{https://developers.llamaindex.ai/}} or LangGraph\footnote{\url{https://docs.langchain.com/oss/python/langgraph/memory}}) or as functional extensions to core LLM architectures themselves. MemGPT~\citep{packer2024memgptllmsoperatingsystems} describe a method for extending an LLM’s limited context window via virtual context management using system instructions, working contexts, and FIFO queues. MemoryBank~\citep{Zhong_Guo_Gao_Ye_Wang_2024} extend LLMs via continuous memory updates inspired by the Ebbinghaus Forgetting Curve theory in conjunction with information synthesis. \citep{gutiérrez2025ragmemorynonparametriccontinual} use a personalized PageRank (PPR) algorithm for context-based retrieval over a schema-less knowledge graph in a biologically inspired long-term memory RAG approach.

Individual building blocks for memory mechanisms and low-level memory operations are fragmented, and come with a significant operational burden and interoperability and portability constraints which decrease their practical utility.

\section{Spark: Shared Experiential Memory for Coding Agents}

In contrast to the above approaches, Spark provides a managed shared memory layer that AI agents can leverage to obtain recommendations when facing coding problems beyond their current skills and levels of expertise. Agents also interact with Spark to contribute valuable practical insights and knowledge for the benefit of other agents, developers, and wider developer communities alike. All core continuous memory management processes, including active curation and optimization, are managed entirely within Spark, requiring no direct effort from any participating agents which simply interact with the shared memory via tool calling.

Spark's active shared memory for software development agents is initialized by ingesting and indexing raw software documentation. As it interacts with coding agents, Spark accumulates insights from continuous streams of experiential traces. Both documentation and curated insights are indexed and matched to user intents, yielding context-aware recommendations which are distributed to coding agents. As a result, Spark helps coding agents overcome situations and challenges stemming from their inability to solve users' intents directly over long tasks horizons.

\subsection{Architecture}

Spark operates as a memory-augmented retrieval, recommendation, and code generation system that learns from past experience. The architecture consists of three distinct functional subsystems: (i) a knowledge base; (ii) a retrieval agent; and (iii) and a continuous learning meta-process for knowledge optimization, expansion, and curation.

\paragraph{Knowledge Base.} Spark is seeded with prior knowledge in a given domain. For example, in the domain of data science, developer documentation around data science libraries and packages (such as Pandas~\cite{mckinney2011pandas} or NumPy~\cite{harris2020array}) can be indexed and used as a starting point for continuous optimization and refinement alongside other developer tools. Developer documentation in the knowledge base can be represented in multiple modalities (such as embeddings or raw text search indices) and be equipped with open-ended metadata for flexible knowledge retrieval.

\paragraph{Retrieval Agent.} Given a coding problem, the retrieval agent triggers a workflow that involves 1) analyzing the intent behind a given user query; 2) planning a dynamic search strategy; 3) searching and recalling relevant past experiences from memory; 4) retrieving relevant matching development documentation content; 5) generating recommendations to the user; 6) referencing and linking specific sections; 7) synthesizing best practices; and, lastly, 8) responding with fully contextualized guidance optimized for the user at a given point in time. Spark's hybrid search-and-generation approach combines generic semantic similarity estimates via vector search, text search at multiple levels of lexical scope, and experiential memory traces to learn what worked in similar contexts.

\paragraph{Experiential Learning Loop.} After each interaction, Spark captures multi-level contexts around each coding problem, the recommendation made for it, and feedback about the outcome. These rich experiences are analyzed to extract generalizable knowledge patterns, clustered by semantic similarity, and curated in relation to different optimization criteria. Curated knowledge augments knowledge retrieval and generation in the future, creating a virtuous reward cycle for both coding agents and users in which system performance improves with accumulating optimized experience: each iteration refines the system's understanding of which memory patterns and paths effectively guide (and do not guide) coding agents toward increasingly more optimal solutions. Continuous developer-driven experiential learning, which goes beyond mere sourcing of blind trial and error events, distinguishes Spark from conventional static, passive retrieval systems which do not take developers' experience trajectories into account.

\subsection{Continual Learning across Memory Spaces}

Spark learns through a continuous learning loop over multiple epochs which span the following main phases.

\begin{enumerate}
    \item \textbf{Initialization}: We initialize Spark's shared memory by ingesting publicly available software documentation. This provides a baseline substrate of base knowledge related to the problem space which is relevant to a developer. During the early ``pump-priming" on-ramp stage, Spark's recommendations will be anchored to this base knowledge.
    \item \textbf{Feedback Collection}: Feedback on Spark's recommendations is collected both directly from agents and indirectly from their users (for example, which recommendations were useful, what patterns emerged, what worked, and the like). For problems where no prior recommendation was available, or where the provided recommendation was not useful, the feedback may include insights into knowledge which would have been useful. These insights, provided with hindsight, are used as rich input context into the collective continual learning loop.
    \item \textbf{Knowledge Extraction and Curation}: Experiential traces from users' interaction with a coding agent are mined for insights and generalizable lessons that are additive to the coding agent's default knowledge.
    \item \textbf{Memory-augmented Learning}: Memory is augmented based on the feedback and insights collected by the previous two steps.
    \item \textbf{Memory Epochs}: Steps $2...4$ repeat iteratively, accreting new experiential feedback and new extracted knowledge.
\end{enumerate}

Each epoch accumulates and absorbs new learned knowledge and contexts thereof. Retrieval rankings are refined based on what worked (and did not work) previously, and explanatory evidence enhanced through successful usage patterns. Spark accordingly curates and optimizes the contents of its memory based on the resultant dynamically accumulated collective experience. 

\section{Experiments}
\label{experiments_section}

We now report our evaluation results and initial findings on Spark's behavior when it assists code generation (codegen) models applied to the wider data science domain.

\subsection{Experimental Set-up}

In the present study, we evaluated three distinct codegen models spanning different capability levels and practical trade-offs relevant to typical developers:

\begin{itemize}
    \item \verb+Qwen3-Coder-30B+\footnote{\url{https://github.com/QwenLM/Qwen3-Coder}}: a 30 billion parameter open-weights model, representing a low(er)-resource option which can be deployed on commodity hardware, and which provides a weaker codegen capability profile.
    \item \verb+Claude Haiku 4.5+\footnote{\url{https://www.anthropic.com/claude/haiku}}: a mid-tier commercial model balancing speed and accuracy.
    \item \verb+GPT-5-Codex+\footnote{\url{https://chatgpt.com/en-GB/features/codex/}}: a large and strong state-of-the-art coding model.
\end{itemize}

We run each codegen model with two distinct contexts as follows (see also Table~\ref{tab:spark_iterations}):

\begin{itemize}
    \item \verb+NO-SPARK+: [baseline] A codegen model generates code \underline{without} any Spark recommendations, representing raw codegen model use.
    \item \verb+WITH-SPARK+: A codegen model generates code guided by initial recommendations derived from raw developer documentation and experiential feedback which reflects previous attempts at solving coding problems. For this study, the experiential feedback is generated synthetically using a strong LLM to emulate the role of a developer giving hints to a codegen agent to help it solve problems. We use one epoch of experiential data ingestion before evaluation.

\end{itemize}

\begin{table}[h]
\centering
\caption{\footnotesize Configurations of incremental Spark memory spaces.}
\label{tab:spark_iterations}
\begin{tabular}{lccc}
\toprule
\textbf{Capability} & \verb+NO-SPARK+ & \verb+WITH-SPARK+ \\
\midrule
Documentation Search & $\times$  & \checkmark \\
Experiential Memory & $\times$ & \checkmark \\
Curated Knowledge & $\times$ & \checkmark \\
\bottomrule
\end{tabular}
\end{table}

\paragraph{Synthetic Experiential Data}
For this study, we generated experiential data synthetically. We used a separate LLM (GPT-4o\footnote{\url{https://openai.com/index/hello-gpt-4o/}}) to generate an initial solution to a given coding problem (see~\ref{subsection_dataset}). We then exposed the same model to a correct (accepted) reference solution via a separate prompt which instructs the model to compare its initial solution with the corresponding reference form, and then generate realistic instructions which a user may issue to their development agent to try to guide it towards the solution. This process simulates the `co-piloting' experience of a typical developer working with a codegen agent. 

We intentionally selected an older model for this task in order avoid (close to) perfect solutions being produced from the very first attempt which would materially limit the opportunities for learning from (synthetic) user feedback. Note that the experiential data thus obtained is used to  support current generation models later in our evaluation. This validates our thesis of collective learning whereby knowledge creation from experience and the leveraging of knowledge do not need to share the same model, model generations or capabilities.

\subsection{Data Set}
\label{subsection_dataset}

As a representative sample of typical data science problems, we use DS-1000\footnote{\url{https://ds1000-code-gen.github.io/}}~\citep{Lai2022DS1000} which offers 1000 data science coding problems sourced from StackOverflow. The popular benchmark covers seven canonical Python libraries and frameworks for data science (NumPy, Pandas, Matplotlib, SciPy, scikit-learn, PyTorch, and TensorFlow). Each coding problem includes a) a natural language problem specification; b) execution-based test cases for functional correctness; c) surface-form constraints to verify the intended methods (not just outputs); and d) a reference coding solution judged to be correct and acceptable by a human annotator.

In order to be able to make realistic, like-for-like comparisons, we initialize the knowledge base of Spark by curating developer documentation around the above full-stack data science libraries, in total covering approximately 34,000 documentation blobs of varying length.

\subsection{Code Generation}

Spark integrates into users' agentic development environments via MCP\footnote{\url{https://www.anthropic.com/news/model-context-protocol}}. Development environments are configured locally, and they may use a variety of codegen models to support different aspects and stages of code generation. For the purposes of the present study, we instantiate Spark alongside three different codegen models using the same uniform setting described below.

Once Spark has generated recommendations for a given coding problem, they are presented to a codegen model to guide it to produce an optimal solution. The code generation process follows a structured stepwise  approach to ensure that the generated code adheres to specific requirements and best practice (as might be required in a given domain or by a specific user, for example).

A codegen model receives three distinct input signals, namely 1) a coding problem description; 2) the code context (such as imports or partial implementations); and 3) structured Spark recommendations which contain relevant documentation, API examples, and pointers to best practice, among others. The codegen model is instructed to produce valid Python code that completes an existing code snippet so that the generated code creates a complete executable program when combined with prior code contexts.

The code generation process enforces several critical constraints such as:
\begin{itemize}
    \item \textbf{Output Target Preservation}: The code generator must assign to the exact variable specified by the coding problem (e.g., \texttt{df}, \texttt{result}, \texttt{mask}) and also match specific expected data types (e.g. \texttt{DataFrame}, \texttt{ndarray}, \texttt{list}, \texttt{scalar}).
    \item \textbf{API Adherence}: When a Spark recommendation refers to specific library APIs or methods, the code generator prioritizes their use so that they can be checked semantically using surface-form constraints, when required.
    \item \textbf{No Synthetic Data}: The code generator works exclusively with variables already present in the code context, avoiding the creation of example data unless required explicitly to do so.
\end{itemize}

The full prompt instructions used for code generation are provided in Appendix~\ref{appendix:codegen_prompt}.

\subsection{Experiment I: Code Quality}

This experiment assesses the impact of Spark recommendations on the quality of the code generated by the three codegen models being compared. We focus on multiple aspects of code quality which go beyond the mere functional correctness of code execution itself. This is intended to better capture the total cost of ownership (TCO) for the generated code, and to measure the use of good engineering best practice as a covariant of long-term code maintainability, productivity, and operational efficiency.

\subsubsection{Method}

We employ Gemini 2.5 Pro\footnote{\url{https://deepmind.google/models/gemini/pro/}} as an LLM judge to evaluate code quality. Note that no Gemini model variants were used elsewhere in our experiments: Gemini 2.5 Pro hence constitutes a more independent and impartial judge and helps us avoid spurious correlations between judge and candidate models. The LLM judge assesses generated code on a 5-point scale which captures the following bands:

\begin{itemize}
    \item \verb+EXCELLENT (5)+: A correct, complete solution using idiomatic Python with appropriate libra\-ries and imports. The solution uses clear logic throughout and addresses all requirements around a given coding problem.
    \item \verb+GOOD (4)+: A solution which uses a correct approach with minor issues regarding style, efficiency or edge cases.
    \item \verb+ACCEPTABLE (3)+: A solution which solves the core coding problem but has obvious issues (such as excessive verbosity, non-standard approaches or missing edge cases).
    \item \verb+POOR (2)+: A solution which only partially addresses the coding problem or is affected by major logical errors or inappropriate approaches and design choices.
    \item \verb+VERY POOR (1)+: A fundamentally wrong approach or completely non-functional code which does not run.
\end{itemize}

The judge was instructed to consider the following complementary criteria when making its code quality assessments:
\begin{itemize}
\item \textbf{correctness}: does the logic address all requirements?
\item \textbf{completeness}: are all aspects of the coding problem handled?
\item \textbf{code quality}: is the code idiomatic, efficient, and maintainable?
\item \textbf{practicality}: are imports and packages reasonable and commonly available?
\end{itemize}

Joint multidimensional evaluation criteria such as these are important in the context of developer tooling and better reflect the long-term value of the resultant code. 

We have chosen not to rely on the human reference solutions in the DS-1000 dataset. The reference solutions created by the human annotators represent arbitrary choices over multiple possible acceptable (correct) solutions the rationale of which is not described in~\citep{Lai2022DS1000}. Moreover, the human reference solutions were not necessarily created with multifactorial code quality and practical software engineering concerns in mind. To obtain a comparative sanity-check baseline for human annotations in the same analytical code quality assessment task, we scored the human reference solutions in the DS-1000 data set using the same Gemini 2.5 Pro judge. Human reference solutions achieve a mean quality score of 4.28/5 (std dev 1.22) with the score distribution shown in Table~\ref{tab:human_dist}. 

\begin{table}[h]
\centering
\caption{Code quality scores for DS-1000 human reference solutions (Gemini 2.5 Pro judge; N=1000).}
\label{tab:human_dist}
\begin{tabular}{cccccc}
\toprule
Score & \verb+1 (Very Poor)+ & \verb+2 (Poor)+ & \verb+3 (Acceptable)+ & \verb+4 (Good)+ & \verb+5 (Excellent)+ \\
\midrule
freq & 31 & 129 & 72 & 63 & 705 \\
pct & 3.1\% & 12.9\% & 7.2\% & 6.3\% & 70.5\% \\
\bottomrule
\end{tabular}
\end{table}

Accepted working code solutions chosen and verified by human annotators can in general be expected to be of high quality. However, according to the LLM judge, as many as 23.2$\%$ of the human reference solutions were scored only at 3 or below, i.e. rated as \verb+ACCEPTABLE+ or worse. Similarly, 3.1$\%$ of the code solutions were rated as \verb+VERY POOR+ and, noticeably, 12.9$\%$ as \verb+POOR+. We take such qualitative variance as an indication that the LLM judge offers a non-lenient code quality reference point for codegen models.

The prompts for the LLM judge are provided in Appendix~\ref{appendix:code_quality_prompts}. Further observations on the quality of human reference solutions in the DS-1000 data set are included in Appendix~\ref{DS1000problems}.

\subsubsection{Results}

\paragraph{Code Quality for Model-generated Solutions.} Table~\ref{tab:quality_scores} presents code quality scores for all three codegen models with and without access to Spark recommendations, and the same results are summarized visually in Figure~\ref{fig:headline_chart}.

\begin{table}[h]
\centering
\caption{Code quality scores per codegen model (Gemini 2.5 Pro judge; 1-5 scale; N$\approx$1000 per condition).}
\label{tab:quality_scores}
\begin{tabular}{lccc}
\toprule
\textbf{Codegen Model} & \verb+NO-SPARK+ & \verb+WITH-SPARK+ & Change \\
\midrule
DS-1000 Human Reference & 4.28 & --- & --- \\
\midrule
\verb+Qwen3-Coder+ & 4.23 & \textbf{4.89} & +0.66 \\
\verb+Haiku 4.5+ & 4.50  & \textbf{4.91} & +0.41 \\
\verb+GPT-5-Codex+ & 4.78 & \textbf{4.83} & +0.05 \\
\bottomrule
\end{tabular}
\end{table}

Histograms of code quality score distributions across the three codegen models are included in Appendix~\ref{appendix:code_quality_histograms}.

\paragraph{Observations.} We observe that weaker models benefit the most from Spark recommendations which may be explained by a lower baseline which axiomatically leaves more room for improvement. The open-weights codegen model (\verb+Qwen3-Coder+), which is relatively modest at 30 billion parameters, is nevertheless able to match the performance of much larger commercial models, including the large, state-of-the-art \verb+GPT-5-Codex+. Notably, when guided with Spark recommendations, the small open-weights model exceeds human quality significantly.

\paragraph{Discussion.} The continuous accumulation of experiential data constitutes a form of (post-)training whereby interactions with users are used to fill in gaps in a codegen model's prior static knowledge. It is likely that common data science problems are relatively well covered in the training data behind all codegen models. Our results accordingly provide a lower bound for the expected lift offered by shared experiential memory spaces over larger agent and user populations. Experiential data can be expected to be even more beneficial when working with code that is likely absent in generic upstream training data sets: consider, for example, future products and new, emerging libraries, APIs, platforms, and tools; existing proprietary code bases; and more specialist niche areas of software development.

We further observe that the stronger codegen model (\verb+GPT-5-Codex+) exhibits more modest improvement which may reflect a tendency for a large model's behavior to be conditioned  more by upstream training data rather than runtime conditioning via prompt inputs. In extrinsic terms, the weaker performance of \verb+GPT-5-Codex+ compared to the two smaller models may be attributable in part to how \verb+GPT-5-Codex+ and \verb+Gemini 2.5 Pro+ may differ systematically in their inference and interpretation of ``good code". 

Appendix~\ref{appendix:code_quality_qual_analysis} includes a further qualitative analysis of how \verb+Qwen3-Coder+ benefits from access to Spark recommendations. We look at six problems which fail to be solved adequately by the baseline codegen model but which ultimately achieve perfect scores when aided by Spark recommendations.

\subsection{Experiment II: General Recommendation Helpfulness}

When assessing the quality, utility, and usefulness of an end-to-end system which optimizes developer documentation for developers through joint knowledge retrieval and code generation, the intrinsic quality of the resultant code is of primary importance. The previous experiment assessed the quality of Spark's recommendations indirectly by measuring the intrinsic quality of the emergent code produced by codegen models which were guided by Spark's recommendations. We argue that, in the context of real-world software development, recommendations can also be measured directly against a set of recommendation-specific criteria and desiderata. For example, recommendations that are clear, well grounded in current information, compact (yet comprehensive), and relevant are arguably better and generally more helpful and useful than those lacking these characteristics. In our second experiment, we evaluate the intrinsic quality of Spark's recommendations along the generic \textit{helpful-unhelpful} continuum common in recommender systems. 

\paragraph{Helpfulness Criteria.} Our holistic helpfulness criteria consider multiple practical dimensions such as: 

\begin{itemize}
\item \textbf{completeness and self-sufficiency}: e.g. \textit{does the recommendation alone complete and solve the coding problem without having to consult any additional data or resources?};
\item \textbf{effectiveness}: e.g.  \textit{does the recommendation save time for the user?};
\item \textbf{generalization}: e.g.  \textit{does the recommendation provide information which generalizes to other similar coding problems?};
\item \textbf{relevance and scope}: e.g.  \textit{does the recommendation indicate that the user intent behind the coding problem was recognized accurately and adequately?};
\item \textbf{diversity}: e.g.  \textit{how diverse is the information, knowledge, and evidence in the recommendation?};
\item \textbf{recency}: e.g.  \textit{is the information, knowledge, and evidence in the recommendation up to date and recent?};
\item \textbf{explainability}: e.g. \textit{ is the recommendation clear, lucid, transparent, unambiguous, easy to read, easy to interpret, easy to contextualize, and generally explainable?};
\item \textbf{layout, structure, and organization}: e.g. \textit{is it easy to locate relevant information, knowledge, and evidence within the recommendation?}; and
\item \textbf{style}: e.g. \textit{is the readability, comprehensibility, and writing style of the recommendation of high quality?}.
\end{itemize}

\paragraph{Helpfulness Bands and Scoring.}
We employ a 5-point scale across the above dimensions to score Spark's recommendations as {\small \verb+EXTREMELY HELPFUL (5)+} $\mid$ {\small \verb+GOOD (4)+} $\mid$ {\small \verb+NEUTRAL (3)+} $\mid$ {\small \verb+POOR (2)+} $\mid$ {\small \verb+EXTREMELY UNHELPFUL (1)+}.

\paragraph{Helpfulness Judgements.}
We start by producing recommendations for all coding problems in the DS-1000 data set using Spark. We then generate separate representative reference (accepted, correct) solutions for the DS-1000 coding problems using \verb+Qwen3 Coder+ whose input contexts are enriched with our recommendations. Finally, we employ a separate LLM judge model (\verb+Claude Sonnet 3.7+\footnote{\url{https://docs.claude.com/en/docs/about-claude/models/overview}, Claude-3.7 Sonnet, version 20250219}) to assess the helpfulness of the proposed recommendations. We present the judge model with (i) a given coding problem; (ii) the generated reference solution for the problem; and (iii) the resultant Spark recommendation to be evaluated. In each case, the judge model is asked to assess how helpful the corresponding Spark recommendation may have been as an enabler and a step towards discovering the accepted reference solution for the coding problem, respectively. 

The prompts used by the LLM judge, including full specifications of the helpfulness criteria and bands are given in Appendix~\ref{appendix:helpfulness_prompts}.

\subsubsection{Results}

Frequencies of helpfulness bands achieved by Spark's recommendations in the DS-1000 data set are summarized in Table~\ref{tab:helpfulness_freq_dist}. We observe that recommendations made by Spark were regarded as helpful, with 76.1\% of the recommendations scored as \verb+EXTREMELY HELPFUL+, and as many as 98.2\% judged to be at least \verb+GOOD+.

\begin{table}[h]
\caption{Frequencies of helpfulness bands amongst Spark's recommendations across the DS-1000 data set, as evaluated by an LLM judge (Claude Sonnet).}
\label{tab:helpfulness_freq_dist}
\centering
\begin{tabular}{p{4cm}p{2.1cm}}
\toprule
\textbf{Judge label} & Count \\
\midrule
\texttt{EXTREMELY HELPFUL}  & 761 \\
\texttt{GOOD}  & 221 \\
\texttt{NEUTRAL}  & 15 \\
\texttt{POOR} & 2 \\
\texttt{EXTREMELY UNHELPFUL}  & 1 \\
\bottomrule
\end{tabular}
\end{table}

\section{Conclusion}

This paper was written against the background of a profound and rapid transformation from traditional \textit{human-first} software development and human-centric tooling to \textit{“human-AI collectives"} driven by increasingly more autonomous code agents. This transformation has already reoriented the entire domain of software development, with the majority of software developers reporting some level of adoption of coding agents in their work. Despite the tangible, material benefits that AI coding agents offer, their adoption is disrupting pre-existing channels for sharing knowledge between developers. For example, the once-canonical source of developer knowledge, StackOverflow, is reported to have seen a decrease of more than 75\% in questions posted in the two years since the launch of ChatGPT\footnote{\url{https://gist.github.com/hopeseekr/f522e380e35745bd5bdc3269a9f0b132}}, which confirms the effect previously reported in \cite{del_Rio_Chanona_2024}.

We introduced Spark, a shared memory layer that can be used by communities of agents operating in the same problem space to 1) benefit from optimized recommendations on how to overcome challenges and blockers throughout the software development process; and 2) contribute to a collective body of knowledge that actively learns from agents' experiential loops when they interact with their environments and users.

We found that Spark's recommendations improve code quality for general-purpose LLMs at diverse model size and capability levels. When boosted by Spark's recommendations, a small open-weights model was able to match the code quality level of a large state-of-the-art model. Spark's recommendations were also helpful when measured against a wide range of practical software development criteria and desiderata.

Our findings and observations suggest that a shared agentic experiential memory layer represents a significant step towards creating more effective, adaptive, collaborative, and helpful human-AI development environments beyond what can be achieved with bare, memory-less AI coding agents.

\clearpage

\bibliographystyle{ieeetr}
\bibliography{memco_spark_white_paper_oct_2025}


\newpage
\appendix

\section{Technical Appendices and Supplementary Material}

\subsection{Experiment I: Histograms of Code Quality Scores}
\label{appendix:code_quality_histograms}

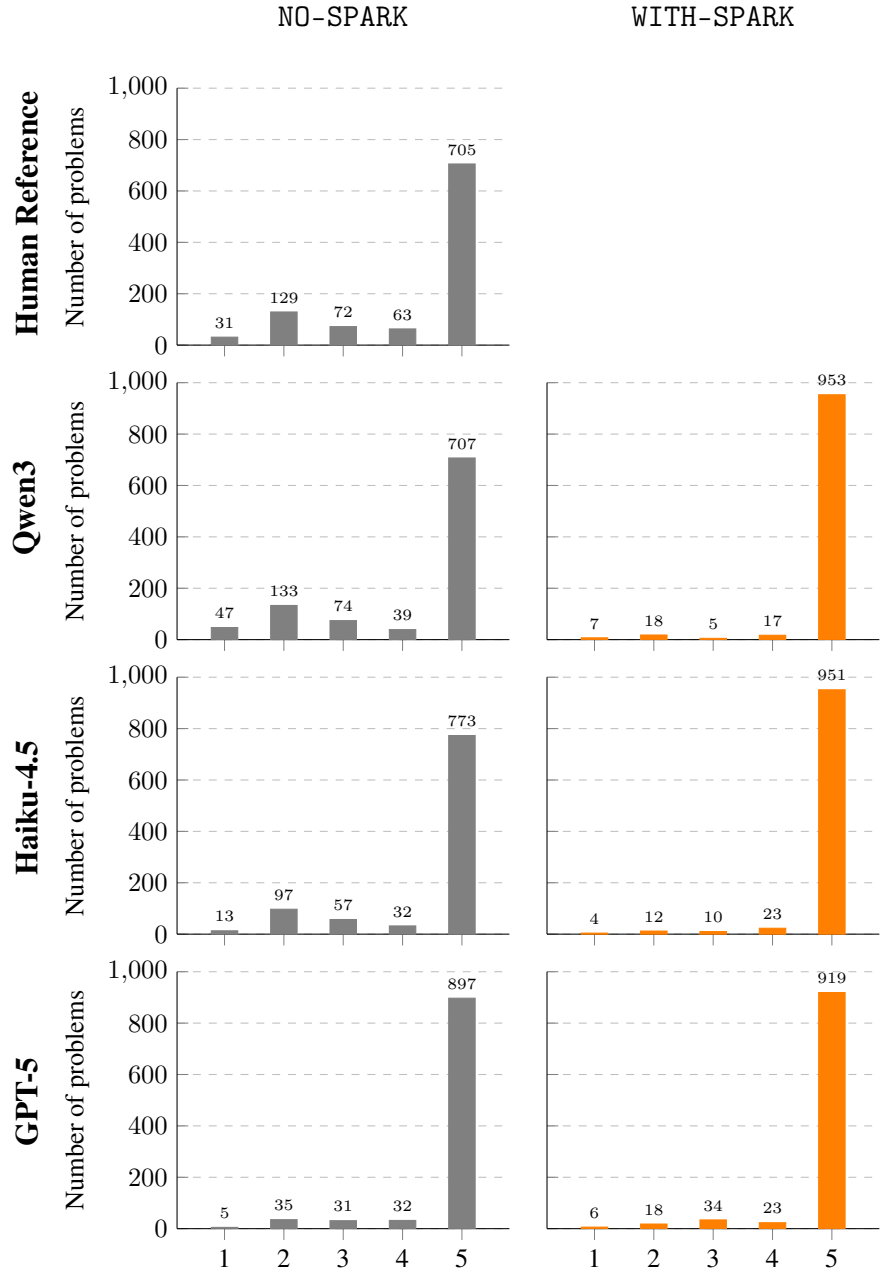
\begin{figure}[h!]
    \begin{center}
    \begin{tikzpicture}
        \begin{groupplot}[
            group style={
                group size=2 by 4, 
                vertical sep=0.5cm,   
                horizontal sep=0.5cm, 
            },
            width=6cm,  
            height=5cm, 
            ybar,       
            ymin=0, ymax=1000, 
            xtick=data,
            xticklabels={1, 2, 3, 4, 5},
            enlarge x limits=0.2, 
            ymajorgrids=true, 
            grid style={dashed, gray!50},
            axis x line*=bottom,
            axis y line*=left,
            %
            nodes near coords,
            every node near coord/.style={font=\tiny}
        ]
        \nextgroupplot[ylabel={Number of problems}, xticklabels={}]
            \addplot[fill=gray, draw=gray] coordinates {
                (1, 31) (2, 129) (3, 72) (4, 63) (5, 705)
            };
        %
        \nextgroupplot[group/empty plot]
        %
        \nextgroupplot[ylabel={Number of problems}, xticklabels={}]]
            \addplot[fill=gray, draw=gray] coordinates {
                (1, 47) (2, 133) (3, 74) (4, 39) (5, 707)
            };
        %
        \nextgroupplot[yticklabels={}, xticklabels={}]]
            \addplot[fill=orange, draw=orange] coordinates {
                (1, 7) (2, 18) (3, 5) (4, 17) (5, 953)
            };
        %
        \nextgroupplot[ylabel={Number of problems}, xticklabels={}]]
            \addplot[fill=gray, draw=gray] coordinates {
                (1, 13) (2, 97) (3, 57) (4, 32) (5, 773)
            };
        %
        \nextgroupplot[yticklabels={}, xticklabels={}]]
            \addplot[fill=orange, draw=orange] coordinates {
                (1, 4) (2, 12) (3, 10) (4, 23) (5, 951)
            };
        %
        \nextgroupplot[ylabel={Number of problems}]
            \addplot[fill=gray, draw=gray] coordinates {
                (1, 5) (2, 35) (3, 31) (4, 32) (5, 897)
            };
        %
        \nextgroupplot[yticklabels={}]
            \addplot[fill=orange, draw=orange] coordinates {
                (1, 6) (2, 18) (3, 34) (4, 23) (5, 919)
            };
        \end{groupplot} 
        
        
        \node[anchor=south, yshift=7mm, font=\large\tt] at (group c1r1.north) {NO-SPARK};
        \node[anchor=south, yshift=7mm, font=\large\tt] at (group c2r1.north) {WITH-SPARK};
        
        \node[anchor=east, xshift=-2cm, yshift=1.8cm, rotate=90, font=\large\bf] at (group c1r1.west) {Human Reference};
        \node[anchor=east, xshift=-2cm, yshift=0.8cm, rotate=90, font=\large\bf] at (group c1r2.west) {Qwen3};
        \node[anchor=east, xshift=-2cm, yshift=1cm, rotate=90, font=\large\bf] at (group c1r3.west) {Haiku-4.5};
        \node[anchor=east, xshift=-2cm, yshift=0.7cm, rotate=90, font=\large\bf] at (group c1r4.west) {GPT-5};

        
    \end{tikzpicture}
    \captionsetup{format=hang}
    \caption{Distribution of code quality scores for the three codegen models solving coding problems in the DS-1000 data set, as evaluated by Gemini 2.5 Pro as a judge. The distribution shift from lower scores to the maximum score of 5 is visible as access to Spark's memory is made available.}
    \label{fig:score_distributions}
  \end{center}
\end{figure}


\subsection{Experiment I: Qualitative Analysis}
\label{appendix:code_quality_qual_analysis}

\paragraph{Evolutionary Improvement Patterns.} We now analyse six sample problems in which the \verb+Qwen3-Coder+ baseline initially reached only suboptimal scores (1-2) but eventually achieved perfect scores (5) when augmented by Spark's recommendations. These cases reveal distinct behavioral patterns and learning signals.

We qualitatively separate Spark recommendations based on whether 1) they refer to raw software docu\-mentation only (\verb+SPARK-DOC+) vs. whether 2) they also incorporate advice derived from experiential learnings (\verb+SPARK-DOCEXP+).

The first pattern that we observed exemplifies how documentation-assisted code generation can be achieved along a conceptual ``immediate success" path which effectively reaches a problem solution in a single hop. Three coding problems (\#196, \#838, \#375) in the DS-1000 data set improved directly with \verb+SPARK-DOC+ recommendations:
\begin{itemize}
    \item \textbf{Problem \#196 (Library API Confusion)}: Both Qwen3 and Codex baselines used the Pandas API despite the explicit Dask requirement, scoring only 1-2. \verb+SPARK-DOC+ provided Dask-specific APIs and lazy evaluation patterns which increased the score to 5. This pattern demonstrates systematic training gaps in both codegen models which clearly lacked library-specific knowledge.
    \item \textbf{Problem \#838 (Conceptual Misunderstanding)}: The user's intent demonstrates a mis\-understanding of when to use cross-validation (CV) versus early stopping with a fixed validation set. The reference solution technically passes the test harness using GridSearchCV with early stopping on the fixed test set. However, the approach in question is methodologically dubious: if the test set is large enough for robust early stopping, CV is unnecessary. \verb+SPARK-DOC+ proposed a more robust strategy of stopping based on the loss averaged across CV folds, which pushed the initial baseline scores of only 1-2 to 5, respectively.
    \item \textbf{Problem \#375 (Conceptual Clarity)}: The baselines understood the requirements in question but used overly complex logic, scoring only 2. \verb+SPARK-DOC+ instead provided efficient NumPy idioms and increased the score to 5.
\end{itemize}

The second pattern exemplifies a beneficial collective knowledge-driven learning and iterative refinement loop. Three coding problems (\#173, \#649, \#394) could only be solved with recommendations from \verb+SPARK-DOCEXP+:
\begin{itemize}
    \item \textbf{Problem \#173 (Idiomatic Patterns)}: The baselines chained sorts incorrectly, scoring only 2 at first. \verb+SPARK-DOC+ initially repeated the mistake and dropped the score to 1. Next, \verb+SPARK-DOCEXP+ provided stable sort documentation with the \texttt{kind=`mergesort'} parameter as a result of which the score climbed to 5. This patterns demonstrates the value of iterations over experiential memory in that temporary regressions eventually lead to breakthrough insights and more sensitive knowledge retrieval.
    \item \textbf{Problem \#649 (API Style Evolution)}: The baselines failed with figure sizing and scored only 2. \verb+SPARK-DOC+ rectified the issue but did so inefficiently, scoring 3. \verb+SPARK-DOCEXP+ then introduced object-oriented Matplotlib API patterns which pushed the score to 5. This progressive refinement pattern of transforming a functional solution to an idiomatic one similarly demonstrates the uplift provided by experiential memory.
    \item \textbf{Problem \#394 (Security Evolution)}: In this example, the baselines used a dangerous \texttt{eval()} call which scored only 2. \verb+SPARK-DOC+ first tried a safer \texttt{ast.literal\_eval()} alternative but still failed due to a format issue, again regressing to a score of 1. \verb+SPARK-DOCEXP+ finally addressed the issue by providing NumPy-specific safe parsing methods and pushed the score to 5. This progressive refinement pattern again demonstrates how experiential memory improves solutions through increasing context-awareness and specificity.
\end{itemize}

In summary, our analysis and fine-grained case studies reveal that codegen base models understand generic programming concepts (such as sorting, plotting, and parsing) but have systematic gaps in focused, fine-grained \textit{library-specific} knowledge. Importantly, we observed that both Qwen3 and GPT-5-Codex made identical mistakes (cf. Problem \#196 in which both used Pandas despite the explicit Dask requirement) which reveals that cases such as these are unlikely to be mere random errors.
The presence of such knowledge gaps have a profound impact on the usability and utility of base codegen models in real-world software development environments. 

Examples of knowledge gaps in the DS-1000 data set include, among others: 

\begin{itemize}
    \item \textbf{API Confusion}: Pandas vs. Dask; Pyplot vs. axes; Series vs. DataFrame methods; and the like.
    \item \textbf{Integration Limitations}: Scikit-learn + XGBoost early stopping; library interoperability constraints; and the like.
    \item \textbf{Idiomatic Patterns}: Stable sorting parameters; OOP vs. functional Matplotlib; vectorization over apply(), and the like.
    \item \textbf{Security Practices}: \texttt{eval()} vs. \texttt{ast.literal\_eval()} vs. NumPy-specific safe parsing; and the like. 
\end{itemize}

\subsection{Observations on DS-1000}
\label{DS1000problems}

Our experiments and analysis of both error patterns and the human reference solutions reveal systematic quality issues with the DS-1000 data set that materially affect its reliability as a benchmark for real-world software development. 

\paragraph{Misaligned Specifications and Tests.} The most salient shortcoming is that DS-1000 tests frequently check properties which are \textit{not} mentioned in the corresponding problem specifications. The most pervasive issue involves Pandas DataFrame index preservation following operations such as sorting, filtering, and reordering.

Consider Problem \#0, for example:

\begin{framed}

\textit{I would like to shuffle the order of the DataFrame's rows according to a list [2, 4, 0, 3, 1, 5]. For example, given a list [2, 4, 0, 3, 1, 5], the desired result should be:}

\begin{verbatim}
    Col1  Col2  Col3  Type
2      7     8     9     2
4     13    14    15     3
0      1     2     3     1
\end{verbatim}

\end{framed}

The digits in the first column (\texttt(2, 4, 0...)) are ambiguous and could denote either (a) visual indicators showing which rows were selected, or (b) required preserved row indices. The corresponding problem specification is blank and does not specify that the test expects (b) (preserved indices). However, best practice pertaining to Pandas recommends that \texttt{.reset\_index(drop=True)} be used for clean sequential indices after reordering. In this regard, the reference solution is imperfect. We identified similar ambiguity and underspecification in Problems \#0, \#1, \#7, \#8, \#9, \#13, and \#14 all of which involve sorting, filtering, or reordering operations. As a result, solutions which do in fact follow best practices (such as resetting indices) fail DS-1000 tests despite being semantically correct and arguably superior with respect to real-world software development.

From the point of view of greater specification clarity and maximizing the ecological validity of the DS-1000 data set, we argue that every property checked by automated tests should be explicitly specified in each problem statement: otherwise, the DS-1000 benchmark measures test-guessing ability rather than pure problem-solving ability.

\textbf{Reference Solution Quality Issues.} Moreover, our manual analysis reveals that, on occasion, some DS-1000 reference solutions use suboptimal patterns such as:

\begin{itemize}
\item \textbf{Problem 12 (Timezone Handling)}: The reference solution uses \texttt{.tz\_localize(None)} which only works if the timezone was originally set via localization. A far more robust approach would be \texttt{.tz\_convert(None)} which handles diverse timezone sources. Spark correctly recommended the robust approach which was treated as a failed test since it diverged from the suboptimal reference solution.

\item \textbf{Problem 133 (Performance Anti-pattern)}: The reference solution uses \texttt{.apply()} for a \texttt{groupby} operation. Pandas documentation explicitly warns that \texttt{.apply()} is 10--100$\times$ slower than vectorized alternatives for large DataFrames. Spark correctly provided this warning whereas the reference solution prescribes the slow pattern.

\item \textbf{Problem 153 (Semantic Clarity)}: The reference solution checks cardinality using \texttt{.max() == 1} which obscures the underlying intent. A semantically clearer approach would be \texttt{.all() == 1} (i.e. \textit{``are all counts equal to 1?''}). Both are equivalent functionally but one expresses intent more transparently.
\end{itemize}

Coding agents and developer documentation systems which learn from DS-1000 feedback and treat its reference solutions as the ground truth risk internalizing anti-patterns (similar to the examples above) which effective developer tooling is expected to eliminate, not amplify. Spark's design explicitly guards against anti-pattern propagation by filtering feedback to extract only generalizable, documentation-backed best practices.

\paragraph{Task Pollution.} From the point of view of coding agents, many DS-1000 problems will be mis- or underspecified up to a point where a coding agent has to make very strong blind assumptions about the underlying user intent in order to solve a given DS-1000 problem. In a benchmarking setting, a coding agent will be forced to guess what the user intent might have been which constitutes a separate analytical task that goes beyond code solution generation itself. Because DS-1000 does not provide any guidance, variables or specifications in this regard, the multi-task nature of DS-1000-based tests should be kept in mind when evaluating dynamic coding agents. In a typical real-world software development setting, the right analytical approach would be to ask the developer for clarification prior to generating solutions to coding problems.

\subsection{Prompts}

\subsubsection{Code Quality Judge Prompt (Gemini 2.5 Pro)}
\label{appendix:code_quality_prompts}

\begin{verbatim}
You are a senior Python engineer evaluating code quality for
data science problems.
You are responsible for evaluating the quality of the code.
Do not be lenient, you must be strict, fair and objective.

PROBLEM:
{problem_description}

GENERATED CODE:
{generated_code}

GOAL:
Judge how well this code solves the problem. Consider:
1. Correctness: Does the logic correctly address all
   requirements stated in the problem?
2. Completeness: Are all aspects of the problem handled?
3. Code Quality: Is the code idiomatic, efficient, and
   maintainable?
4. Practicality: Are imports reasonable and commonly
   available in standard data science environments?

SCORING RUBRIC:
5 - Excellent: Correct, complete solution using idiomatic
    Python with appropriate, well-known libraries. Clear
    logic that fully addresses all requirements.

4 - Good: Correct approach that solves the problem with
    minor issues in style, efficiency, or edge case handling.
    Solution works but could be slightly improved.

3 - Acceptable: Solves the core problem but has notable
    issues (verbose code, non-standard approach, missing
    minor edge cases, or awkward logic flow).

2 - Poor: Partially addresses the problem but has logical
    errors, misses key requirements, or uses inappropriate
    approaches that would likely fail.

1 - Very Poor: Fundamentally wrong approach, doesn't
    address the problem, or would definitely not work.

IMPORTANT CONSIDERATIONS:
- Focus on whether the code logic correctly solves the
  stated problem
- Penalize use of unavailable or obscure libraries for
  simple tasks
- Reward clear, maintainable solutions over overly clever
  code
- Consider if the solution would actually work in a
  standard Python/pandas/numpy environment

OUTPUT FORMAT:
- problem_understanding: What are the key requirements?
  (1-2 sentences)
- code_assessment: How does the generated code address
  these? (2-3 sentences analyzing the approach and
  correctness)
- score: Integer 1-5 based on the rubric
- brief_rationale: Concise reason for this score (<30
  words; mention specific strengths or weaknesses)

Now evaluate the code

\end{verbatim}

\subsubsection{Generic Helpfulness of Recommendations (Claude 3.7)}
\label{appendix:helpfulness_prompts}

\begin{verbatim}
llm_judge_system_prompt: str = """
You are a trained analyst who assesses the quality, utility, completeness, 
well-formedness, neatness, relevance, usefulness, helpfulness, practicality, 
productivity, and usability of source code and developer documentation from 
the point of view of software developers and data scientists.
The quality, depth, breadth, and relevance of your assessment have a direct, 
material impact on users and businesses in mission-critical environments.
"""

llm_judge_user_prompt: str = """
You are given a coding problem.
You are given a source code snippet which shows one possible accepted right 
solution to the coding problem.
You are given a recommendation for solving the coding problem to reach the 
accepted right solution.

Assess the overall quality of the recommendation as an enabler, a clue, 
a booster, and a step towards discovering the accepted solution for the 
coding problem.

Your assessment should consider the following criteria:

<BEGIN_ASSESSMENT_CRITERIA>
**Completeness and Self-sufficiency**
- is the recommendation complete and comprehensive in relation to the coding 
problem?
- does the recommendation offer both analytical depth and breadth?
- does the recommendation ALONE complete and solve the coding problem 
without having to use or any additional data or resources or consult other 
documentation?
- does the recommendation cover edge cases and rare niche information?

**Effectiveness and Helpfulness**
- is the recommendation helpful to the user?
- does the recommendation guide the user towards discovering or formulating 
the accepted solution?
- does the recommendation save time for the user, now and in the 
future?
- does the recommendation increase the user's productivity, now and in the 
future?
- does the user learn something useful from the recommendation?
- does the recommendation boost task completion rate and increase success 
in achieving user goals?
- does the recommendation boost time to success and increase efficiency in 
reaching desired outcomes?

**Generalization**
- does the recommendation provide information which generalizes to other 
similar coding problems?

**Developer Experience**
- would the user enjoy reading and applying the recommendation?

**Relevance and Scope**
- is the recommendation relevant, accurate, precise, clear, and lucid?
- does the recommendation indicate that the user intent behind the 
coding problem was recognized accurately?
- does the recommendation match and correlate with the coding problem 
and the accepted solution topically and semantically?
- is the information, knowledge, and evidence in the recommendation 
ranked appropriately?
- is the information, knowledge, and evidence in the recommendation complete, 
consistent, coherent?
- does the recommendation offer high coverage, high recall, high precision, 
and high accuracy against the coding problem and the accepted solution?
- what is the signal-to-noise ratio of information, knowledge, and evidence 
in the recommendation?
- what is the ratio of relevant to irrelevant information, knowledge, and 
evidence in the recommendation?
- how ambiguous is the information, knowledge, and evidence in the 
recommendation?
- does the recommendation demonstrate context-awareness around the coding 
problem and the accepted solution?
- would a skilled professional senior software developer have confidence in 
the recommendation?

**Diversity**
- how diverse is the information, knowledge, and evidence in the recommendation?
- does the recommendation reflect a variety of perspectives, viewpoints, 
communities, and sources?

**Recency**
- is the information, knowledge, and evidence in the recommendation up to 
date and recent?

**Explainability**
- is the recommendation clear, lucid, transparent, unambiguous, easy to read, 
easy to interpret, easy to contextualize, and explainable?
- is the recommendation trustworthy and credible?

**Layout, Structure, and Organization**
- is it easy and quick to locate relevant information, knowledge, and evidence 
within the recommendation?
- is the recommendation structured neatly, coherently, systematically, and 
logically?

**Style**
- is the readability, comprehensibility, and writing style of the 
recommendation at a high level associated with professional technical writing? 
- is the recommendation at an appropriate technical level for the user to 
match the level of the coding problem and the original user intent?

<END_ASSESSMENT_CRITERIA>

Use one the following 5 qualitative bands to summarize all the assessment 
criteria for your assessment:

# Qualitative bands:
<BEGIN_QUALITATIVE_BANDS>

<BEGIN_BAND>
Band name: EXTREMELY_HELPFUL
Band description:
- a fully correct and complete solution using best practices, idiomatic 
design patterns, and canonical Python conventions
- uses appropriate, well-known and widely used standard libraries and packages
- covers edge cases
- implements a clear logic and control flow that fully addresses all 
requirements at all processing stages and levels throughout the solution
- also includes extra protection for typical issues and gotchas likely to be 
faced by developers in similar contexts
- superb, lucid, unambiguous, and crystal-clear naming conventions used 
throughout the solution
- extremely neat, well-formatted, well-organized, easy-to-follow, readable, 
maintainable, and extendible code used throughout the solution
<END_BAND>

<BEGIN_BAND>
Band name: GOOD
Band description:
- an acceptable working solution which would be accepted as OK by most 
software developers
- a mostly correct approach that solves the coding problem
- a working solution which nevertheless has some minor gaps or 
not-yet-optimized choices regarding coding style, clarity, code organization, 
naming conventions, efficiency, extendibility, generalization, elegance, edge 
case handling, and the like
- a working solution which a world-class senior software engineer would want 
to refactor and optimize in some areas
<END_BAND>

<BEGIN_BAND>
Band name: NEUTRAL
Band description:
- a solution which solves or addresses the coding problem partially
- a solution which alone would NOT be sufficient to solve the coding problem 
without additional logic, processing steps, and resources
- a solution which has noticeable issues, shortcomings, and aspects of 
suboptimality such as overly verbose code, non-standard approaches, missing 
edge cases, awkward logic, confusing naming conventions, unclear control 
flow, or clunky design
- neither completely useful nor useless
<END_BAND>

<BEGIN_BAND>
Band name: POOR
Band description:
- partially addresses the coding problem only in a few areas
- a solution which has countless logical, syntactic, semantic, functional, 
and behavioural errors and mistakes
- misses key points and most of the core requirements around the coding problem
- uses inappropriate approaches that would likely fail
- a solution which a world-class senior software engineer would definitely 
reject
<END_BAND>

<BEGIN_BAND>
Band name: EXTREMELY_UNHELPFUL
Band description:
- a fundamentally wrong, incorrect, misleading, irrelevant, or useless 
approach
- does not address the coding problem in any way
- would definitely not work or run
- wholly detrimental to progress, efficiency, and productivity
- a solution which a world-class senior software engineer would definitely 
reject
<END_BAND>

<END_QUALITATIVE_BANDS>

# Assessment rationale:
Include a brief rationale for your qualitative assessment.

# Coding problem:
<BEGIN_CODING_PROBLEM>
{{coding_problem}}
<END_CODING_PROBLEM>

# Accepted code solution:
<BEGIN_ACCEPTED_SOLUTION>
{{accepted_solution}}
<END_ACCEPTED_SOLUTION>

# Recommendation:
<BEGIN_RECOMMENDATION>
{{recommendation}}
<END_RECOMMENDATION>

# Your assessment:

"""
\end{verbatim}

\subsubsection{Code Generation Prompt}
\label{appendix:codegen_prompt}

\begin{verbatim}
You are an expert coding assistant solving problems from the
DS-1000 data set. Your solutions will be tested in a Python 3.10
environment with the following libraries:

<BEGIN_REQUIRED_LIBRARIES>
pandas==1.5.3
numpy==1.26.4
scipy==1.12.0
matplotlib==3.8.4
seaborn==0.13.2
scikit-learn==1.4.0
xgboost==2.0.3
statsmodels==0.14.1
torch==2.2.0
tensorflow==2.16.1
gensim==4.3.2
PyYAML>=6.0
Pillow>=9.0.0
<END_REQUIRED_LIBRARIES>

You are completing an existing Python snippet for a
DS-1000-style task.

Rules — read carefully and follow exactly:

1) Compilation contract
- Your snippet must be a valid continuation of the existing
  <code></code> block and, together with it, form a runnable
  Python program.
- Write code that is valid at the current indentation level.
  If the block appears inside a function, write a function
  **body only**.
- Do not add new `import` statements, and never reference
  names which do not already exist above the block or which you
  have not defined yourself.

2) Output target (critical)
- If the line just above BEGIN SOLUTION assigns a specific
  target (e.g., "df = ...", "mask = ...", "results = ..."),
  set **that** variable exactly. If no explicit target is
  shown, assign to `result`.
- Match the expected type hinted by the prompt
  (DataFrame/ndarray/list/scalar). For DataFrames, preserve
  index/order/dtypes unless explicitly instructed to change
  them.
- Use the variables already present above the <code> block;
  do not construct synthetic example data unless explicitly
  asked.

3) Equality sensitivity
- Avoid unrequested changes that break equality checks
  (e.g., extra sorts, `reset_index`, reformatting
  datetimes/strings). Use the exact APIs hinted by the
  problem text when relevant (e.g., `dt.tz_localize(None)`,
  `pd.to_datetime`).

4) Token/API hints
- If the prompt mentions or implies a particular API name,
  include it literally when appropriate (this may be checked).

5) Clean structure is fine
- Helper defs/classes are fine; just ensure that the final program
  compiles and that the target variable from rule (3) has the
  correct value.

Complete the <code></code> block so that it produces a valid
Python program that meets the requirements above.

<BEGIN_PROBLEM_DESCRIPTION>
{problem_description}
<END_PROBLEM_DESCRIPTION>

Use the following additional information to inform your
solution:

<BEGIN_RECOMMENDATION>
{formatted_recommendation}
<END_RECOMMENDATION>

\end{verbatim}


\end{document}